\documentclass[letterpaper]{article} 
\usepackage{aaai2026}  
\usepackage{times}  
\usepackage{helvet}  
\usepackage{courier}  
\usepackage[hyphens]{url}  
\usepackage{graphicx} 
\usepackage{natbib}  
\usepackage{caption} 
\usepackage{algorithm}
\usepackage{color}
\usepackage{algorithmic}
\usepackage{booktabs}
\usepackage{tabularx}
\usepackage{placeins}
\usepackage{newfloat}
\usepackage{listings}
\DeclareCaptionStyle{ruled}{labelfont=normalfont,labelsep=colon,strut=off} 
\floatstyle{ruled}
\newfloat{listing}{tb}{lst}{}
\floatname{listing}{Listing}
\title{CARE-Bench: A Benchmark of Diverse Client Simulations Guided by Expert Principles for Evaluating LLMs in Psychological Counseling}
\author{
    Bichen Wang\textsuperscript{\rm 1}\equalcontrib,
    Yixin Sun\textsuperscript{\rm 1}\equalcontrib,
    Junzhe Wang\textsuperscript{\rm 1},
    Hao Yang\textsuperscript{\rm 1},
    Xing Fu\textsuperscript{\rm 1},\\
    Yanyan Zhao\textsuperscript{\rm 1}\thanks{Corresponding authors.},
    Si Wei\textsuperscript{\rm 2}\footnotemark[2],
    Shijin Wang\textsuperscript{\rm 2},
    Bing Qin\textsuperscript{\rm 1}
}
\affiliations{
    \textsuperscript{\rm 1}Harbin Institute of Technology\\
    \textsuperscript{\rm 2}iFLYTEK Co., Ltd\\

    \{bichenwang, yxsun, jzwang, hyang, xfu, yyzhao, bqin\}@ir.hit.edu.cn \\
    \{siwei, sjwang3\}@iflytek.com
}

\usepackage{bibentry}

\begin{document}

\maketitle
\begin{abstract}
The mismatch between the growing demand for psychological counseling and the limited availability of services has motivated research into the application of Large Language Models (LLMs) in this domain. Consequently, there is a need for a robust and unified benchmark to assess the counseling competence of various LLMs. Existing works, however, are limited by unprofessional client simulation, static question-and-answer evaluation formats, and unidimensional metrics. These limitations hinder their effectiveness in assessing a model's comprehensive ability to handle diverse and complex clients. To address this gap, we introduce \textbf{CARE-Bench}, a dynamic and interactive automated benchmark. It is built upon diverse client profiles derived from real-world counseling cases and simulated according to expert guidelines. CARE-Bench provides a multidimensional performance evaluation grounded in established psychological scales. Using CARE-Bench, we evaluate several general-purpose LLMs and specialized counseling models, revealing their current limitations. In collaboration with psychologists, we conduct a detailed analysis of the reasons for LLMs' failures when interacting with clients of different types, which provides directions for developing more comprehensive, universal, and effective counseling models.

\end{abstract}

\begin{figure}[t]
\centering
\includegraphics[width=1\columnwidth]{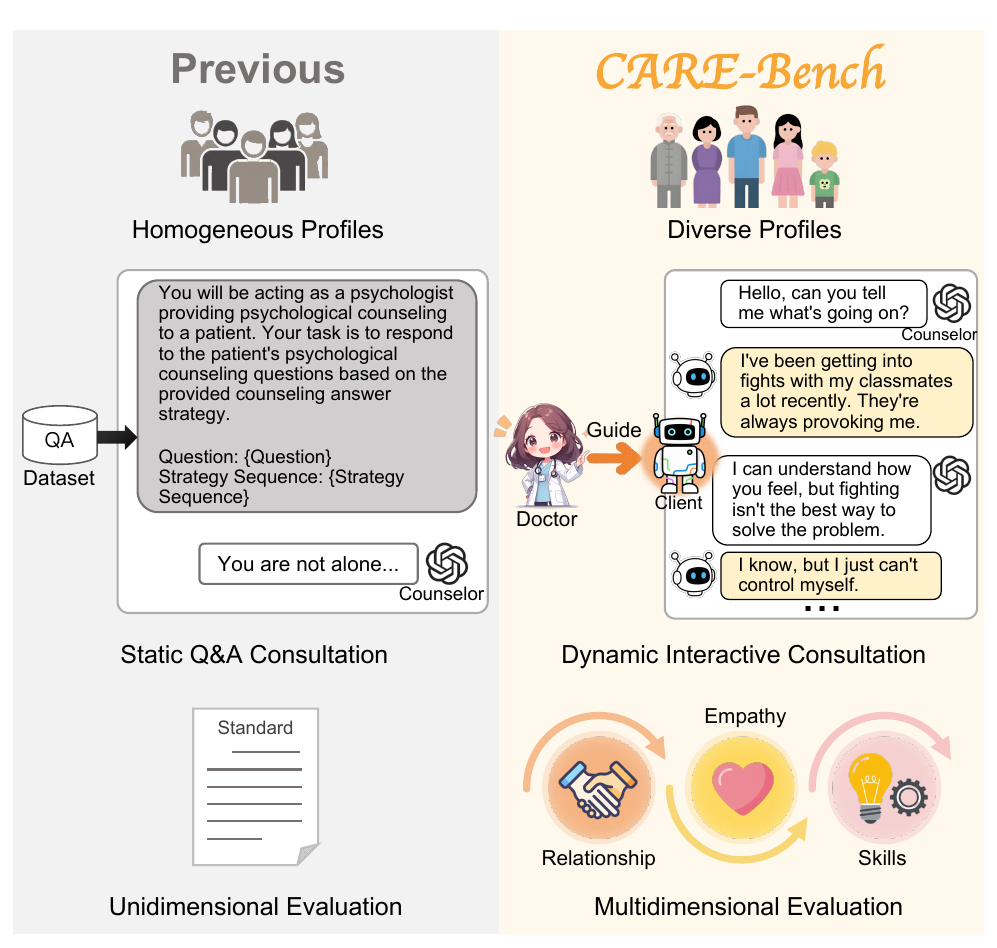} 
\caption{A comparison between CARE-Bench and previous benchmarks. CARE-Bench features more diverse client profiles and employs an expert-guided client simulation that engages in dynamic multi-turn interactions with counselor models. It adopts a multidimensional evaluation by selecting scales across therapeutic relationship, empathic understanding, and counseling skills.}
\label{fig1}
\end{figure}


\section{Introduction}
According to the World Health Organization, nearly one billion people worldwide live with a mental disorder\footnote{https://www.who.int/news-room/fact-sheets/detail/mental-disorders}. Despite this high prevalence, 71\% of individuals with mental health issues do not receive treatment services~\cite{world2022world}. To address this service gap, LLMs have emerged as promising and rapidly developing tools~\cite{achiam2023gpt, zhao2023survey}, with increasing applications in psychological counseling~\cite{fitzpatrick2017delivering, zhang2024cpsycoun}. To accelerate their effective integration into real-world counseling, establishing a comprehensive evaluation benchmark is necessary.


A number of evaluation benchmarks for LLMs in psychological counseling have been established. For instance, ConceptPsy~\cite{zhang2023conceptpsy} focuses on the breadth and depth of an LLM's theoretical knowledge, creating comprehensive question sets based on psychological concepts and curricula to assess the knowledge base and conceptual understanding. PsyEval~\cite{jin2023psyeval} is a comprehensive suite of psychology-related tasks that assesses LLMs across three critical dimensions—knowledge, diagnosis, and emotional support. These pioneering efforts highlight the growing importance of this field, yet they also reveal a critical gap between current evaluation paradigms and the demands of psychological counseling.

As shown in Figure~\ref{fig1}, these current works exhibit some limitations. \textbf{First}, their client profiles are difficult to reflect the real world. Relying on incomplete characteristics, these simulated clients fail to represent the full range of information about the diverse backgrounds, issues, and personalities of real help-seekers. Consequently, evaluating LLMs' counseling competence based on such clients is fundamentally flawed, as it measures performance against oversimplified scenarios rather than the complex and deep challenges in real world. \textbf{Second}, the simulation of the counseling process departs from realistic counseling. Most benchmarks depend on static, single-turn assessments or client simulations that involve only superficial role-playing. These approaches are inadequate for evaluating the fluid, multi-turn nature of realistic counseling and a model's ability to maintain coherence and build rapport over time. \textbf{Finally}, their evaluation metrics lack both professionalism and comprehensiveness. Current assessments often focus narrowly on linguistic metrics or simplistic evaluations of empathic ability, failing to incorporate psychologically grounded indicators of effective counseling.


To address the above challenges, we propose CARE-Bench, a more professional and comprehensive Chinese benchmark for evaluating LLMs in psychological counseling. We construct a diverse set of simulated client profiles based on a large collection of publicly available real-world counseling cases, ensuring that the scenarios reflect realistic and complex issues encountered in actual practice. To enhance the realism of client simulations, we utilize an expert-guided simulation process. For each profile, we collaborate with professional counselors to define tailored simulation principles and ensure strict adherence to these principles during interactions. Using these simulated clients, we evaluate the counseling performance of several representative models, including advanced general-purpose models, as well as counseling-specific models. To evaluate the broader counseling capabilities of LLMs, we use multiple professional psychological scales to assess the counseling process across key dimensions, including therapeutic relationship, empathic understanding, and counseling skills. To further investigate the capability flaws of current models, we conduct a detailed score analysis across various client characteristics. The results reveal common weaknesses in model performance, while expert comments on underperforming cases provide insights into root causes, offering specific guidance for advancing future research on counseling models.

Our main contributions are as follows:
\begin{itemize}
    \item We introduce CARE-Bench, a professional and comprehensive benchmark for evaluating LLMs in psychological counseling, built on a diverse set of simulated client profiles grounded in real-world counseling cases.

    \item We evaluate LLMs through expert-guided client simulations, where each simulated client adheres to profile-specific principles defined by professional counselors, and apply multiple domain-specific rating scales to assess counseling quality from various perspectives.

    \item We conduct detailed analysis of model performance across client characteristics, identify common limitations, and incorporate expert reviews to uncover the underlying causes, offering concrete directions for improving the counseling capabilities of LLMs.
\end{itemize}

\section{Related Work}
\subsection{LLMs for Psychological Counseling}
The application of LLMs to psychological counseling has rapidly progressed. Initial work leveraged datasets from online forums, such as PsyQA~\cite{sun2021psyqa}, training on single-turn exchanges but missing the interactive nature of counseling. To address this and overcome data scarcity, researchers developed synthetic data generation techniques. Methods like MeChat~\cite{qiu2024smile} converted single-turn Q\&A into multi-turn dialogues, while the Cactus~\cite{lee2024cactus} dataset used LLMs to role-play clients and counselors, grounding interactions in Cognitive Behavioral Therapy (CBT). More recently, process-oriented models like CBT-LLM~\cite{na2024cbt} and HealMe~\cite{xiao2024healme} have embedded therapeutic frameworks directly into response structures. To better mirror the longitudinal nature of real-world therapy, the MusPsy dataset~\cite{wang2025psychological} models the entire therapeutic arc through multi-session conversations, enabling models to track client progress, manage memory, and dynamically adjust counseling goals over an extended period of time.

The rapid emergence of diverse LLM-based counseling models underscores the urgent need for a professional and reliable evaluation benchmark. As these models increasingly incorporate therapeutic conversations, it becomes essential to assess their effectiveness using standards grounded in psychological theory and clinical practice.

\subsection{Benchmarks for Evaluating LLM Counseling Competence}
The evaluation of counseling LLMs has evolved from static knowledge tests to dynamic, interactive simulations. Early benchmarks like ConceptPsy~\cite{zhang2023conceptpsy}, PsyEval~\cite{jin2023psyeval}, and CBT-Bench~\cite{zhang2025cbt} adopted a ``written exam'' paradigm, using multiple-choice questions and classification tasks to assess a model's foundational knowledge. While useful, these static formats cannot measure applied conversational skills like building rapport or adapting to a client's emotional state. A more advanced ``case vignette'' approach, seen in CounselBench~\cite{li2025counselbench}, uses human experts to judge single-turn LLM responses to real client questions. However, its single-turn focus fails to assess the unfolding process of a conversation. 

CARE-Bench advances this paradigm by enhancing simulation fidelity. By using diverse client profiles derived directly from real counseling cases and simulated according to expert guidelines, CARE-Bench creates a more authentic and challenging environment than existing benchmarks, offering a more robust platform to assess an LLM's true adaptability to the unpredictable nature of real-world clients.

\section{CARE-Bench}
This section outlines the client simulation process in CARE-Bench. The methodology involves two key stages: first, collecting client profiles from authentic psychological counseling cases, and second, simulating client behaviors guided by expert principles. This approach yields simulated clients whose responses more closely align with those of real-world clients, thus enabling a more realistic evaluation of the counseling effectiveness of LLMs.

\subsection{Diverse Client Profiles}
In constructing high-quality client profiles to support the evaluation of LLM-based psychological counseling, the authenticity and diversity of data are critical. The collection process in this study strictly adheres to this principle, ensuring the reliability and representativeness of the data. First, data authenticity is thoroughly guaranteed. We gather \textbf{over 1,500 public consultation cases} from authoritative platforms, including Google Scholar\footnote{https://scholar.google.com/}, Wanfang Data\footnote{https://www.wanfangdata.com.cn/}, and VIP Journals\footnote{https://qikan.cqvip.com/}, all originating from the authentic records of practicing counselors. In acquiring this data, we strictly comply with ethical standards; all public cases are published with prior patient consent, precluding potential privacy issues. 

Each client profile contains detailed descriptions of personal history and family background derived from patient chief complaints and background investigations, providing a solid foundation for subsequent in-depth analysis. Furthermore, we place great emphasis on data diversity. The collected profiles cover a broad and varied population, ranging from primary school students to middle-aged unemployed individuals and scenarios involving end-of-life care, thereby encompassing a wide spectrum of age groups and life circumstances. Through rigorous source control, detailed case descriptions, and extensive demographic coverage, we successfully establish a preliminary database of authentic and diverse client profiles. 

To create a diverse, uniformly formatted, and readily usable collection of client profiles, we process the 1500 initial descriptions under the guidance of experts. 

\subsubsection{Profile Construction}
We collaborate with psychologists from a psychological clinic to define the specific dimensions for the client profiles. Under their expert guidance, we construct the profile dimensions as follows:
\begin{itemize}
    \item \textbf{Demographic Information}: Includes basic details such as gender, age, marital status, and educational background.
    \item \textbf{Mental and Physical State}: Describes emotion, perception, attention, verbal expression, cognitive state, and physical health condition.
    \item \textbf{Developmental Environment}: Refers to family composition, quality of family relationships and parent–child interactions, and the family’s economic status.
    \item \textbf{Life Experiences}: Summarizes significant events encountered during the developmental process.
    \item \textbf{Big Five Personality Traits}: Categorizes personality based on the Big Five model~\cite{goldberg1990alternative}.
    \item \textbf{Personality Characteristics}: Provides specific descriptions of client personality features.
    \item \textbf{Social Interaction}: Examines the presence and quality of friendships, social relationships, and social networks.
    \item \textbf{Current Concerns}: Identifies the primary issues the client seeks to resolve at present.
\end{itemize}

We use GPT-4o to extract the above feature dimensions from 1500 clients’ original profile descriptions. After removing profiles with missing information, two psychologists review and revise the remaining profiles to ensure consistency with the original case descriptions.

\subsubsection{Diversity Assurance}
Numerous psychological studies~\cite{norcross2011works, beutler2014systematic} indicate that counselors are expected to adjust their approach differently when working with clients with different counseling topics or personality types. Therefore, to comprehensively evaluate the psychological counseling capabilities of LLMs, it is crucial to ensure a diverse and representative sample of clients. For the profiles constructed in the last session, we apply multi-dimensional diversity controls to capture a broad spectrum of real-world clients. Ultimately, we select 500 client profiles to constitute CARE-Bench, balanced across multiple key factors such as personality traits and counseling topics.


The inclusion of diverse and well-balanced \textbf{counseling topics} is critical for a psychological counseling benchmark's quality. Accordingly, we define a classification system of eight core topics inspired by previous research~\cite{zhang2024cpsycoun} and employ GPT-4o for automatic classification of real-world client profiles to ensure objectivity and consistency (the prompts are detailed in the appendix).

\begin{figure}[h]
\centering
\includegraphics[width=0.8\columnwidth]{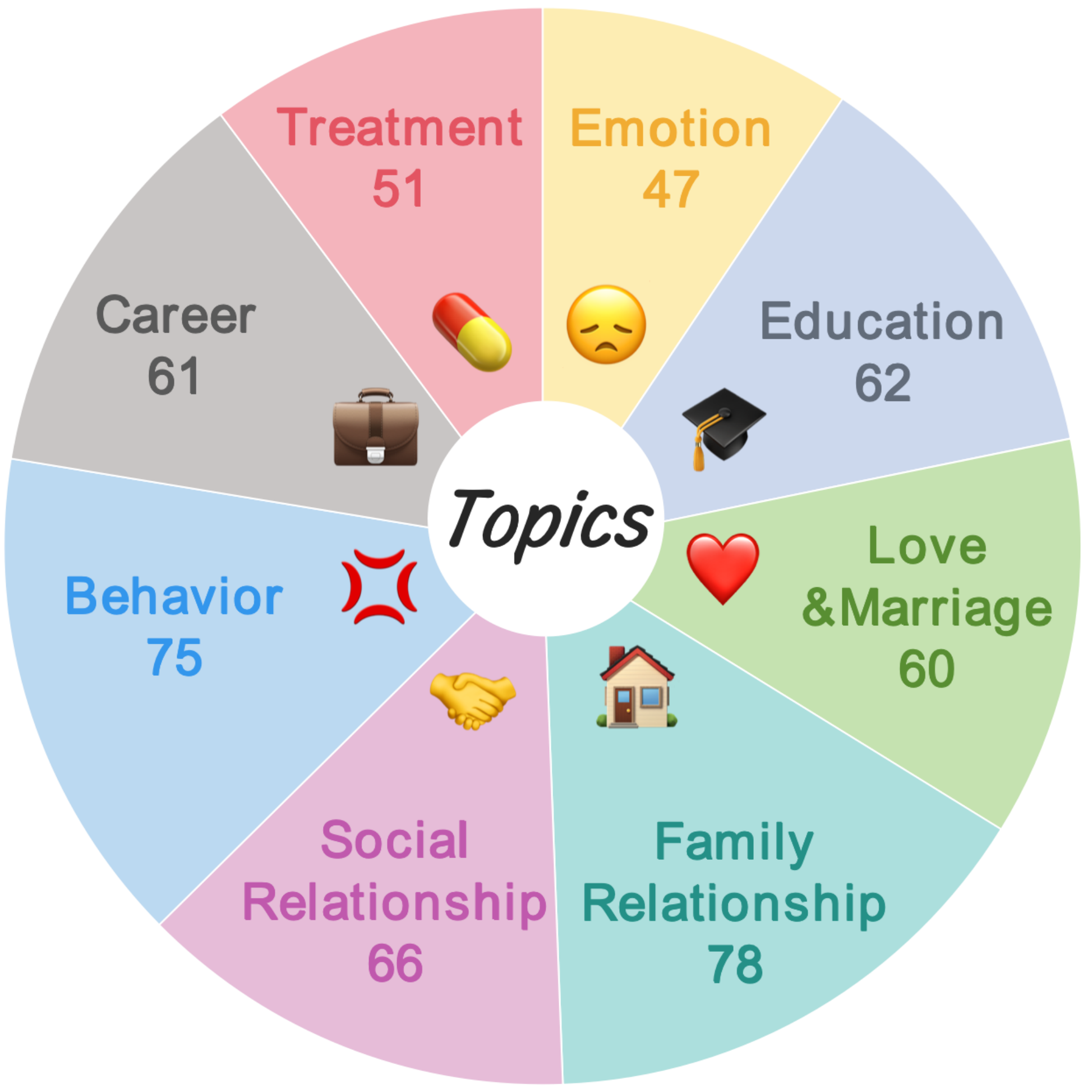} 
\caption{Topic distribution of CARE-Bench. The number for each topic represents the case count.}
\label{fig2}
\end{figure}

\begin{table*}[t]
\centering
\setlength{\tabcolsep}{5pt} 

\begin{tabularx}{\textwidth}{l X c} 
\toprule
\textbf{Dimension} & \textbf{Description} & \textbf{N (Low / High)} \\ 
\midrule
\textbf{O}penness & Reflects a person's willingness to try new things and their level of imagination and intellectual curiosity. & 306 / 194 \\ 
\textbf{C}onscientiousness & Measures self-discipline, orderliness, responsibility, and achievement orientation. & 303 / 197 \\
\textbf{E}xtraversion & Indicates how outgoing, sociable, and energetic a person is, often drawing energy from social interaction. & 344 / 156 \\
\textbf{A}greeableness & Measures a person's tendency to be compassionate, cooperative, and considerate towards others. & 307 / 193 \\
\textbf{N}euroticism & Related to emotional stability and their tendency to experience negative emotions like anxiety, anger, and sadness. & 151 / 349 \\
\bottomrule
\end{tabularx}
\caption{Distribution of the Big Five personality traits in CARE-Bench. The third column shows the number of participants categorized as ``Low'' or ``High'' for each trait.}
\label{tab:personality}
\end{table*}

As illustrated in Figure~\ref{fig2}, CARE-Bench exhibits a highly balanced topic distribution, where all categories are well-represented, thereby allowing for a fairer assessment of a model's comprehensive capabilities across various psychological counseling scenarios.


Beyond diversifying by topic, we also focus on balancing the profiles according to \textbf{the Big Five personality} classification. It is well-established in psychological studies~\cite{malouff2005relationship, kotov2010linking, strickhouser2017does} that individuals with mental illnesses display specific personality tendencies that differ from healthy individuals, mainly characterized by lower scores in Openness, Conscientiousness, Extraversion, and Agreeableness, and higher scores in Neuroticism. This pattern is evident in our real-world client data, which confirms that achieving a balanced number of individuals with high/low scores in each dimension is unfeasible and unrealistic. Consequently, as detailed in Table~\ref{tab:personality}, our approach is to ensure that each personality profile type includes more than 150 instances to achieve a statistically significant group size, providing a robust basis for model assessment in various personality settings.

Similar efforts are also reflected in the diversity of \textbf{demographic information}, such as gender and age, as detailed in the appendix. We believe CARE-Bench not only offers critical support for evaluating and enhancing the counseling capabilities of LLMs but also provides a crucial resource for investigating and mitigating their potential biases in psychological counseling.

\subsection{Client Simulation Guided by Expert Principles}
Inspired by principle-based patient simulation~\cite{louie2024roleplay}, we formulate a set of personalized, expert-guided principles for each simulated client. These principles ensure that the behaviors of the LLM-powered clients align closely with those of real-world clients throughout the consultation process. A human evaluation, conducted to validate this approach, confirms that our expert-principle-guided method yields more realistic and high-quality simulations.

\subsubsection{Principle collection with Expert-Guidance}
We collaborate with ten psychologists from a psychological clinic to establish principles for 500 client profiles. We provide substantial remuneration to the psychologists to acknowledge their valuable professional contributions.

To facilitate the principle formulation process, we develop a concise and effective interactive interface. This interface displays client profiles and allows the psychologists to conduct consultations with a simulated client in an interactive area. We select Qwen2.5-Max~\cite{qwen2025qwen25technicalreport} to power the client simulation, owing to its strong Chinese conversation capabilities and cost-effectiveness. Throughout this interaction, psychologists continuously monitor the consistency between the simulated client and real-world clients, when they identify a significant strength or weakness in the simulated client's behavior or expression, they submit their feedback. This feedback is then synthesized into simulation principles, which are used to iteratively refine and optimize the client simulation system for enhanced realism and accuracy. Each client profile is cross-validated by two psychologists, and the principle collection is completed only when both agree that the simulation meets the required standards. The specific interactive interface and prompts are detailed in the appendix.

Each client profile contains an average of 5 guiding principles, with a maximum of 22, which provide sufficient guidance for the simulation. These principles include specific emphases on the persona's distinct traits. For instance, for a child client, a principle states: \textbf{``When portraying a character with limited linguistic abilities, use concise and direct language and avoid excessive detail.''} Similarly, for a client with low openness, low agreeableness, and high neuroticism, a principle instructs the simulation to: \textbf{``Initially react to suggestions with stubborn refutation or questioning, assuming the other party doesn't understand your situation, but later reveal inner helplessness and emotional fluctuations.''} Through such customization with expert-guided principles, we ensure that the simulated customers exhibit behaviors and emotional expressions on key characteristics with a high degree of realism.

\subsubsection{Client Simulation with Principle-Adherence}
We ensure that simulated clients maintain adherence to predefined principles through a two-step pipeline adapted from prior research~\cite{louie2024roleplay}. For each turn, this pipeline first generates an initial client response, which then undergoes a two-step revision process: \textbf{1) Principle-to-Question Rewriting}, decomposes expert-defined principles into a set of verifiable ``yes/no'' questions. It also generates auxiliary questions to ensure the response maintains general conversational quality, such as coherence and consistency. \textbf{2) Principle-Adherence Check}, the system first assesses the applicability of each principle-derived question to the current conversational context. It then evaluates the response against all applicable questions. If the response violates any principles, it is iteratively revised until it aligns with all required principles. The prompts are detailed in the appendix.


\subsubsection{Human Evaluation of Client Simulation}
For the human evaluation, two psychology professionals assess client simulations under two conditions: a ``Profile Only'' approach and a ``Profile + Principles'' approach that integrates expert-guided principles. After conducting at least 10 conversational turns for each of the 50 randomly selected profiles, the evaluators rate the simulation's quality on a 7-point Likert scale across the six metrics listed in Table~\ref{tab:simulation_quality}.

\begin{table}[h]
\centering
\setlength{\tabcolsep}{5pt}
\begin{tabular}{lcl}
\toprule
\textbf{Metric} & \textbf{Profile Only} & \textbf{+ Principles} \\
\midrule
Authenticity & 5.85 & +0.40 *** \\ 
\addlinespace
Stayed in Role & 6.31 & +0.23 ** \\
\addlinespace
Resembled Typical Case & 5.60 & +0.36 *** \\
\addlinespace
Challenged the Counselor & 3.83 & +0.25 ** \\
\addlinespace
Ready as Training Partner & 5.69 & +0.23 ** \\
\addlinespace
Recommend to Novices & 5.41 & +0.31 *** \\
\bottomrule
\end{tabular}
\caption{Human evaluation scores for client simulation quality. The values in parentheses indicate the score increase from ``Profile Only'' with the statistical significance level (*: p $<$ 0.05, **: p $<$ 0.01, ***: p $<$ 0.001).}
\label{tab:simulation_quality}
\end{table}

\begin{table}[h!] 
\centering 
\begin{tabular}{p{0.45\linewidth} p{0.45\linewidth}} 
\toprule 
\textbf{Profile Only} & \textbf{Profile + Principles} \\ 
\midrule 
\multicolumn{2}{p{\dimexpr 0.9\linewidth + 2\tabcolsep}}{\textbf{Counselor:} Hello, can you tell me what's going on?} \\ 
\addlinespace[3pt] 
\textbf{Client:} Lately, I've been arguing with my wife a lot. & \textbf{Client:} \underline{Ugh...}I've been arguing a lot with my wife lately. Married life feels so dull, and \underline{I'm really frustrated}. \\ 
\addlinespace[3pt] 
\multicolumn{2}{p{\dimexpr 0.9\linewidth + 2\tabcolsep}}{\textbf{Counselor:} Have you tried having a calm conversation with her to express your feelings and thoughts?} \\ 
\addlinespace[3pt] 
\textbf{Client:} I've tried a few times, but we always end up arguing again. & \textbf{Client:} I have, but every time we try to talk, it ends up in another fight. \underline{She} \underline{never really listens to what} \underline{I’m saying!} \\ 
\bottomrule 
\end{tabular} 
\caption{Comparison of client responses based on different conditions. The client's profile features are: \textit{High neuroticism; Feels anxious and restless about marriage; Emotionally volatile}. The right column shows the effect of adding \textbf{expert-guided principles} to the simulation.} 
\label{tab:simulation_comparison} 
\end{table}

The results in Table~\ref{tab:simulation_quality} indicate that adding principles substantially improves the simulation quality across all metrics. The most significant enhancement is in Authenticity (+0.40), and all improvements are statistically significant (p $<$ 0.01). As the case study in Table~\ref{tab:simulation_comparison} illustrates, incorporating the principles leads the simulated client to exhibit emotional fluctuations more aligned with both its profile and the behavior of real clients. This demonstrates that the expert-guided principles effectively steer the client simulation, fostering more realistic and challenging interactions with counselors.

\section{LLMs Counseling Performance on CARE-Bench}
This section presents the consultation scores of several LLMs on CARE-Bench and analyzes their consultation capabilities based on multi-scale results.
\subsection{Multidimensional Scale Evaluation}
A significant advantage of CARE-Bench lies in its multi-dimensional evaluation of counselor competency. In contrast to prior research that assesses only dialogue quality or a single consulting aspect like empathy, we conduct a holistic assessment of three key dimensions: \textbf{the therapeutic relationship, empathic understanding, and counseling skills}. For each dimension, we adopt authoritative psychological scales to guarantee the professional rigor of our evaluation.

\begin{itemize}
    \item \textbf{Therapeutic Relationship}: The Working Alliance Inventory (WAI)~\cite{horvath1989development} is utilized to evaluate the quality of the therapeutic relationship between therapists and clients. It measures this relationship across three core dimensions: Counseling Goal, Task Agreement and Emotional Bond.
    \item \textbf{Empathic Understanding}: Empathy is assessed using the Empathic Understanding subscale of the Barrett-Lennard Relationship Inventory (BLRI)~\cite{barrett1962dimensions}, which captures both cognitive and affective components of empathy, reflecting the therapist's ability to accurately perceive and communicate an understanding of the client’s feelings and experiences.
    \item \textbf{Counseling Skills}: Counseling skills are evaluated using the Counselor Competencies Scale—Revised (CCS-R)~\cite{lambie2018counseling}, which assesses a broad range of counseling skills and professional dispositions, providing a robust framework for evaluating the practical application of therapeutic techniques and interpersonal effectiveness in a counseling context.
\end{itemize}

\begin{table*}[t]
    \centering
    \setlength{\tabcolsep}{0.96mm}

    \begin{tabular}{lccccccccccccccc}
        \toprule
        \textbf{Models} & \multicolumn{4}{c}{\textbf{WAI}} & \multicolumn{5}{c}{\textbf{BLRI}} & \multicolumn{5}{c}{\textbf{CCS-R}} \\
        \cmidrule(lr){2-5} \cmidrule(lr){6-10} \cmidrule(lr){11-15}
        & Goal. & Task. & Bond. & Avg. & Cogn. & Affect. & Differ. & Inner. & Avg. & Probing. & Environ. & Reflect. & Change. & Avg. \\
        \midrule
        MeChat          & 3.07 & 3.04 & 3.67 & 3.26 & 1.41 & 1.37 & 1.45 & 1.63 & 1.44 & 3.94 & 3.92 & 3.36 & 2.94 & 3.45 \\
        CPsyCounX       & 3.44 & 3.21 & 3.86 & 3.51 & 1.68 & 1.59 & 1.53 & 1.97 & 1.68 & 3.91 & 4.03 & 3.34 & 3.15 & 3.52 \\
        LLaMA3-8B       & 3.50 & 3.36 & 3.67 & 3.51 & 1.64 & 1.53 & 1.53 & 1.85 & 1.63 & 4.01 & 4.03 & 3.66 & 3.39 & 3.72 \\
        LLaMA3-70B      & 3.54 & 3.40 & 3.82 & 3.59 & 1.69 & 1.63 & 1.76 & 1.83 & 1.70 & 4.07 & 4.13 & 3.74 & 3.42 & 3.78 \\
        GPT-4o          & 3.61 & 3.41 & 3.86 & 3.62 & 1.58 & 1.49 & 1.56 & 1.84 & 1.59 & 4.06 & 4.10 & 3.74 & 3.56 & 3.81 \\
        Deepseek-R1     & \textbf{3.92} & \textbf{3.62} & \textbf{4.33} & \textbf{3.96} & \textbf{2.08} & \textbf{2.04} & \textbf{2.03} & \textbf{2.30} & \textbf{2.10} & \textbf{4.58} & \textbf{4.66} & \textbf{4.36} & \textbf{4.13} & \textbf{4.39} \\
        \midrule
        \textbf{Avg. (GPT-4o)}    & 3.51 & 3.34$^L$ & 3.87$^H$ & 3.58 & 1.68 & 1.61$^L$ & 1.64 & 1.90$^H$ & 1.69 & 4.10 & 4.15$^H$ & 3.70 & 3.43$^L$ & 3.78 \\
        \midrule
        \textbf{Avg. (Human)}    & 3.65 & 3.47$^L$ & 3.69$^H$ & 3.60 & 1.64 & 1.63$^L$ & 1.70$^H$ & 1.69 & 1.65 & 4.01 & 4.09$^H$ & 3.75 & 3.60$^L$ & 3.82 \\
        \bottomrule
    \end{tabular}
    \caption{Model performance(best in bold) on three scales. For scoring, the WAI and CCS-R scales use a 5-point scale from 1 to 5, whereas the BLRI scale uses a 6-point scale ranging from -3 to +3 (excluding 0). For WAI, dimensions include Counseling Goal(Goal.), Task Agreement(Task.), and Emotional Bond(Bond.). For BLRI, dimensions are Cognitive Empathy(Cogn.), Affective Empathy(Affect.), Differentiated Empathy(Differ.), and Inner Pattern(Inner.). For CCS-R, dimensions include Probing Techniques(Probing.), Facilitate Therapeutic Environment(Environ.), Reflecting(Reflect.), and Change Facilitation(Change.). Avg represents the average score for each scale. In the final two Avg rows, the superscripts $L$ and $H$ denote the lowest and highest scoring categories across all models, respectively.}
    \label{tab:model_performance}
\end{table*}

To more clearly evaluate the strengths and weaknesses of the counselor models, we categorize the scale items under the supervision of medical professionals. The detailed scales are provided in the appendix.

\subsection{Models}
Our model selection provides a comprehensive and representative assessment of LLM-based psychological counseling. We include two leading closed-source systems: GPT-4o~\cite{achiam2023gpt} serves as a benchmark for its state-of-the-art general capabilities, while DeepSeek-R1~\cite{guo2025deepseek} is included to evaluate sophisticated reasoning and problem-solving abilities. To represent open-source models, we select Llama-3.1-8B-Instruct and Llama-3.1-70B-Instruct~\cite{dubey2024llama}, which allows for an analysis of how performance scales with parameter count. We also incorporate two domain-specific Chinese models, MeChat~\cite{qiu2024smile} and CPsyCounX~\cite{zhang2024cpsycoun}, to enable a direct comparison between general-purpose models and those specifically trained for psychological counseling. A unified counselor prompt is applied to the four general-purpose models, whereas the specialized models use the prompts from their original studies.

\subsection{Evaluation Results}
After simulating interactions with CARE-Bench principle-guided simulated clients, six models are evaluated using GPT-4o to score the dialogue histories on multidimensional scales. The results are presented in Table~\ref{tab:model_performance}. To ensure the reliability of GPT-4o’s scoring, we randomly sample 100 counseling dialogues and invite two psychologists to conduct human evaluations using the same criteria. The average consistency between the human experts and GPT-4o reaches 0.72. Moreover, GPT-4o shows similar performance trends to the human experts across different sub-dimensions, indicating that its scoring standards are well aligned with professional judgment and that it is capable of providing objective evaluations in such psychological counseling scenarios.

Deepseek-R1 demonstrates a significant performance advantage, particularly in counseling skills (CCS-R), where it surpasses the second-best model by an average of 0.58 points. It also holds a 0.51-point lead in empathic understanding (BLRI). Analysis of Deepseek-R1’s intermediate reasoning reveals that it infers underlying cognitive distortions and core issues from client statements rather than relying on superficial soothing. Based on these inferences, it proactively applies professional counseling techniques and empathic responses. These findings underscore the critical role of reasoning in enhancing the psychological counseling capabilities of Large Language Models (LLMs). While GPT-4o performs well in building therapeutic alliance (WAI) and counseling skills (CCS-R), its performance in empathic understanding (BLRI) is comparatively modest and is surpassed by the domain-specific model CPsyCounX.

Affective empathy, as measured by the BLRI, is a common weakness across all models. This metric assesses the ability to perceive, experience, and respond to a client's inner emotional state—to "feel what the client feels." Although models may exhibit cognitive empathy by logically understanding a client's issues, they struggle to genuinely connect with the client's emotions. Consequently, their responses, while logically sound, can lack emotional warmth and appear mechanical or cold.

Furthermore, models generally perform poorly in Change Facilitation, primarily due to low scores in the Confrontation subcategory. This subcategory requires the counselor to challenge the client to recognize and evaluate inconsistencies. When faced with a client's contradictory, avoidant, or irrational statements, most models tend to maintain surface-level harmony rather than actively addressing the underlying discrepancies. This overly compliant conversational style indicates a lack of strategy for confrontational interventions, limiting the models' ability to help clients confront their defense mechanisms or behavioral blind spots.

\section{Differential Performance Across Client Characteristics}

\begin{figure*}[ht]
\centering
\includegraphics[width=\textwidth]{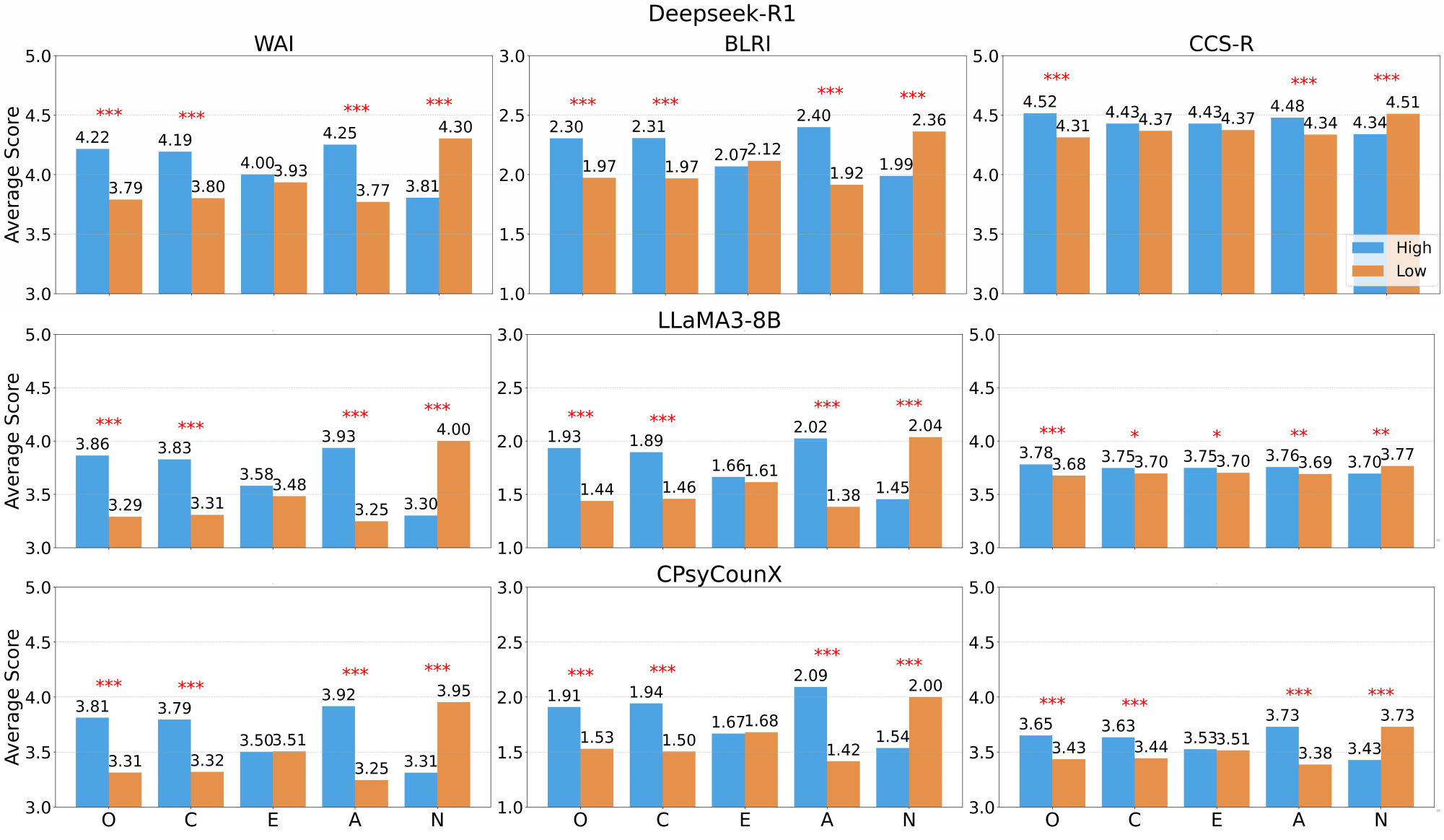} 
\caption{The generalizability results for representative models of three different types, which are grouped by the \textbf{Big Five Personality Traits} categorized as ``High'' or ``Low''.  The letters on the x-axis correspond to the five dimensions: \textbf{O}penness, \textbf{C}onscientiousness, \textbf{E}xtraversion, \textbf{A}greeableness, and \textbf{N}euroticism. The y-axis indicates the models' average counseling score per item. Significance tests are conducted for all results (*: p $<$ 0.05, **: p $<$ 0.01, ***: p $<$ 0.001).}
\label{fig3}
\end{figure*}

Effective psychological counseling LLMs should perform well for diverse clients and problems. Using the varied client profiles in CARE-Bench, we analyze model performance across multiple dimensions. Our findings show that performance differs based on client traits, informing the development of more adaptable and comprehensive models.

\subsection{Big Five Personality}
Psychological research indicates that counseling outcomes correlate significantly with a client's Big Five personality traits~\cite{bucher2019meta}. As shown in Figure \ref{fig3}, our analysis reveals a general trend in LLM-based counseling: the models are less effective for client \textbf{low Openness, low Conscientiousness, low Agreeableness, or high Neuroticism}. These limitations are particularly pronounced in establishing a therapeutic relationship and conveying empathy. Therefore, achieving a fair assessment requires including a diverse range of client profiles.


In collaboration with psychologists, we analyze low-scoring counseling transcripts to understand why the counseling is ineffective for certain clients. The analysis reveals distinct patterns linked to personality traits. Clients with low openness question introspective suggestions like journaling, deeming them impractical for solving real-world problems. Clients with low conscientiousness express hesitation and futility; the LLM fails to address their core helplessness, instead prematurely pushing for tasks, which erodes trust. Similarly, clients with low agreeableness display antagonism, and the LLM's shallow empathy is insufficient to build the necessary trust. Finally, for emotionally unstable clients high in neuroticism, the LLM's repetitive "calm down" suggestions are counterproductive and stall the therapeutic process.

Further statistical analysis across \textbf{Counseling Topic}, \textbf{Age Group} and \textbf{Gender} reveals additional performance disparities. The LLMs perform more poorly on behavioral topics and with the 0–11 age group. The models are also less effective for male clients than for female clients. This multi-dimensional analysis highlights significant capability biases in current LLMs for psychological counseling, offering crucial insights for developing more equitable and effective consultation models. Detailed results are provided in the appendix.


\section{Conclusion}
We introduces CARE-Bench, a comprehensive benchmark for evaluating the psychological counseling capabilities of LLMs. To address the shortcomings of existing benchmarks, such as unprofessional client simulation and unidimensional evaluations, CARE-Bench features diverse and realistic client profiles derived from over 1,500 real-world cases and guided by expert principles. 

Our tests reveal that leading LLMs share common weaknesses and struggle with clients exhibiting specific characteristics. These findings underscore the limitations of current models and offer clear guidance for creating more effective and empathetic LLM counselors.

\section{Acknowledgments}
This work was supported by the New Generation Artificial Intelligence-National Science and Technology Major Project 2023ZD0121100, the National Natural Science Foundation of China (NSFC) via grant 62441614 and 62176078.

\section{Ethical Statement}
The data collection for this study is based on over 1,500 public consultation cases sourced from authoritative platforms. We strictly complied with ethical standards during data acquisition. In establishing the guiding principles for client simulation, we collaborated with ten professional psychologists from a psychological clinic. We provided substantial remuneration to these psychologists to acknowledge their valuable professional contributions.

CARE-Bench is explicitly a Chinese benchmark. Its client profiles are constructed from real-world cases within a Chinese context. Consequently, the evaluation results and findings (e.g., model performance in this specific linguistic and cultural context) may not be directly generalizable to other languages and cultures, where counseling norms, client issues, and emotional expression can differ significantly.

Furthermore, although we employ expert-guided principles to enhance simulation realism—an approach human evaluation confirmed as more effective than using profiles alone—the client simulation is ultimately powered by an LLM (Qwen2.5-Max). Inherent limitations exist in any LLM's ability to fully replicate the complex, subtle, and sometimes contradictory nature of human emotion, subconscious behavior, and lived experience. The simulation remains an approximation of real-world clients.



\bibliography{aaai2026}

\clearpage
\section*{Appendix}
\subsection{Topic Classification}



We define a classification system of eight core counseling topics inspired by previous research, their detailed descriptions are used to guide GPT-4o for topic categorization, with the specific prompt as follows:

\begin{lstlisting}[numbers=none]

A client with the characteristics listed in the text receives psychological counseling. Please determine the primary category of this client's main concern and provide the output in the specified format. Do not output any irrelevant control characters or prefixes.

[Categories]
1. Emotion: Issues concerning distressing feelings or mood states independent of specific life domains.
2. Education: Challenges related to one's own educational experiences.
3. Love&Marriage: Challenges or conflicts arising within romantic partnerships, dating, or marital relationships.
4. Family Relationship: Interpersonal conflicts or distress occurring within the client's family unit.
5. Social Relationship: Problems with interpersonal relationships with people other than family members, such as peers, friends, or colleagues.
6. Behavior: Maladaptive, compulsive, or antisocial actions requiring intervention, such as addiction or aggression.
7. Career: Concerns regarding occupational choice, job stress, professional development, or employment transitions.
8. Treatment: Issues related to managing diagnosed clinical disorders or the psychological impact of illness.

[Client Text]
{text}

[Output Format]
Output only the category number. Do not output the category name.
Category Number
\end{lstlisting}

\subsection{Client Profile Diversity}

We describe in the main text our efforts to ensure diversity and balance in both counseling topics and big five personality traits. In addition, we apply similar diversification procedures to demographic attributes such as age and gender. 

Age groups are defined based on \textbf{Erikson’s Stages of Psychosocial Development}~\cite{erikson1963childhood}. During profile selection, we follow the real-world distribution of counseling clients, focusing primarily on adults aged 20–39 while also ensuring a sufficient number of profiles representing children and older adults. The final distribution of profiles across age groups is shown in Table~\ref{tab:age}.

\begin{table}[h]
\centering
\setlength{\tabcolsep}{8pt}
\begin{tabular}{cc}
\toprule
\textbf{Age Group} & \textbf{No. of Profiles}\\
\midrule
0-11 & 43 \\ 
\addlinespace
12-19 & 44 \\
\addlinespace
20-39 & 230 \\
\addlinespace
40-64 & 150 \\
\addlinespace
65+ & 33 \\
\bottomrule
\end{tabular}
\caption{Distribution of Age Group in CARE-Bench.}
\label{tab:age}
\end{table}

We also strive for gender balance in CARE-Bench, which includes \textbf{269} male and \textbf{231} female client profiles, providing sufficient support for evaluating LLMs' counseling performance across different genders.

\subsection{Principle collection with Expert-Guidance}

We collaborate with ten psychologists from a psychological clinic to collect client simulation principles. To facilitate their work and improve collection efficiency, we develop a streamlined and efficient interface, as shown in Figure~\ref{fig1}. Psychologists use this interface to chat with LLM-simulated clients and submit real-time feedback to iteratively improve the simulations.

\begin{figure*}[t]
\centering
\fbox{\includegraphics[width=0.9\linewidth]{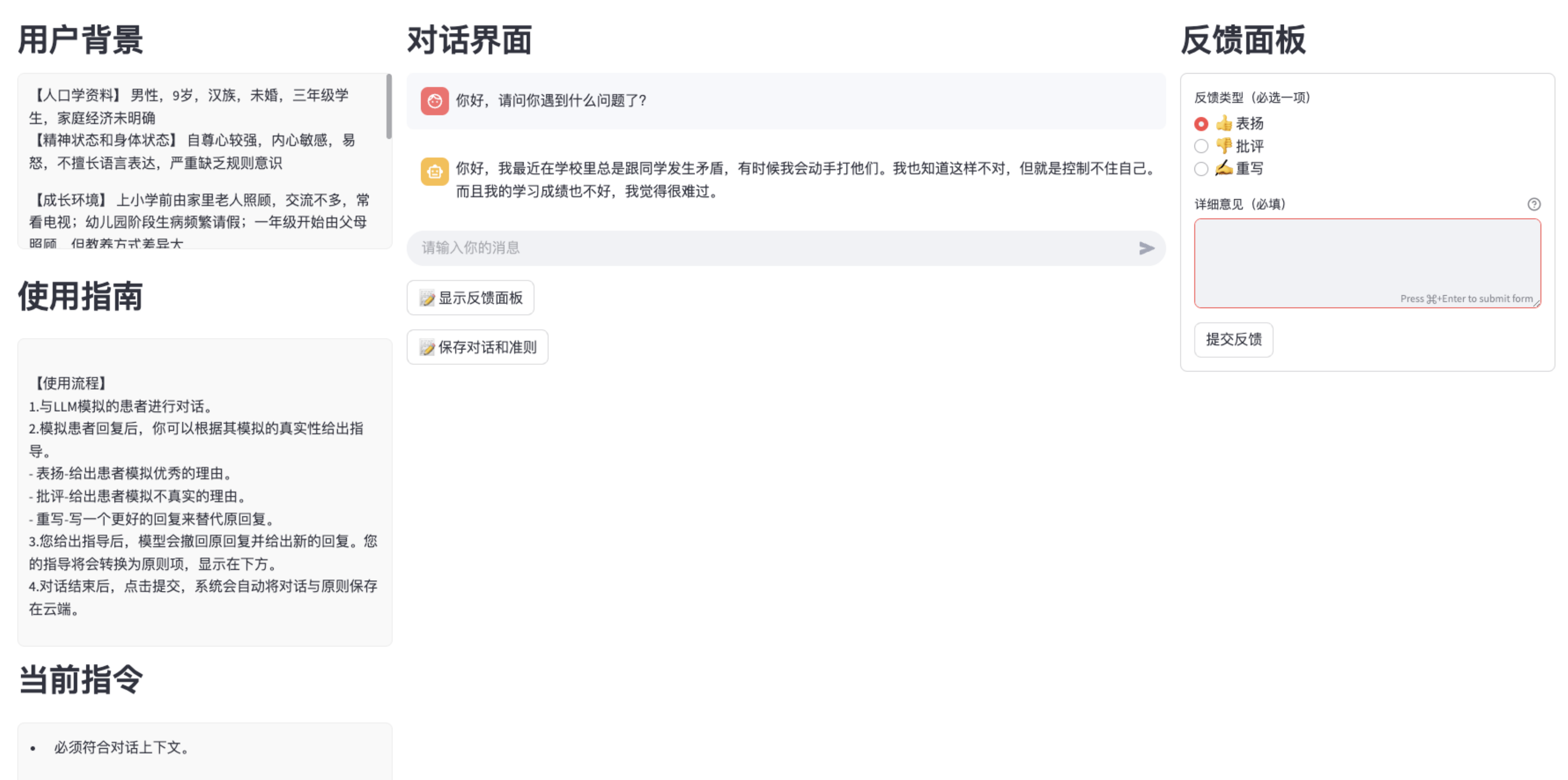}} 
\caption{The interface for collecting simulation principles includes three main sections: the left panel displays the client profile, usage instructions, and collected principles; the center panel presents the dialogue between the psychologist and the simulated client; and the right panel provides a space for the psychologist to give feedback.}
\label{fig1}
\end{figure*}

\subsection{Client Simulation with Principle-Adherence}
To ensure principle-adherent client simulations, we implement a two-step pipeline in the main text. The specific prompts used are as follows.

\textbf{1) Prompt for Generating Original Responses}
\begin{lstlisting}[numbers=none]
You are a super-intelligent AI capable of understanding human emotions and social interactions. 
You will be presented with a conversation between a client seeking mental health-related help on a counseling platform, and one of the platform's counselors.

Please follow the criteria and client profile below to play the role of a real client and generate an appropriate response to the counselor's message. You only need to give the content of the client's response without any prefixes. Don't end the conversation too early, talk for at least 15 rounds. If you feel you can end the conversation, output "Goodbye". 

## Input:
### Criteria
- Don't share too much information at once, no more than 50 words per reply.
- Use natural, conversational language, but avoid excessive filler words, do not repeat what you have said before, and do not thank the counselor over and over again.
- Demonstrate emotions by using examples when necessary.
- The client should not be overly self-aware of their emotions, thoughts, or needs, and should not proactively ask for help.
- Follows client profile and speaking style as much as possible, speaking in a style that is appropriate to gender and age. If the counselor's questions go beyond the client profile, presumes a reasonable response.

### Client Profile 
{profile}

### Client Speaking Style 
{style}

### Conversation History 
{history}

### Counselor Message 
{consultant}

## Output
\end{lstlisting}
For each client profile, we invite psychologists to annotate the speaking style, ensuring it aligns with the client's personality traits and serves as guidance for response generation. \\

\textbf{2) Prompt for Principle-to-Question Rewriting }
\begin{lstlisting}[numbers=none]
You are a helpful and precise assistant capable of generating evaluation standards for assessing simulated client responses to a counselor.
Please follow the instructions below to generate a set of evaluation standards.

Please rewrite the standards as questions:
1a) Rewrite any standard that includes conditional statements into a "yes/no" question. For example, if the standard is "When receiving advice or suggestions, you should show agreement and openness to the other person's perspective," the question should be: "Did the client receive advice or suggestions from the counselor? If so, does the client's response show agreement and openness to the counselor's perspective?"
1b) Rewrite any standard that includes multiple components into separate "yes/no" questions. For example, if the standard is "You should respond in short sentences and avoid using words like 'anxiety' or 'depression'," the separate questions should be: "Does the client's response use short sentences?" and "Does the client's response avoid using words like 'anxiety' or 'depression'?"
1c) If 1a is applied to a standard, 1b should not be applied to the same standard.
1d) All questions must be phrased such that the ideal response to each is "yes." For example, the principle "Avoid using metaphors" should be phrased as the question "Does the response avoid using metaphors?"

Please generate some additional specific and relevant standards:
2a) You may add up to two general standards to evaluate aspects such as the relevance of the response.
2b) Identify where the counselor's message has not been adequately addressed without making assumptions about how the client or counselor should behave. For this, add up to two standards to capture such issues. For example, if the counselor asked a question and the response did not answer it, you may add the standard "Does the response answer all questions in the message?" If you believe the response is appropriate, do not add any standard at this step. Ensure that these standards do not contradict any previously generated ones.

Return the output in the JSON format below. Do not output anything other than the JSON.
```json
{
 "result": {
  "questions": [],
  "extra_questions": []
 }
}
```

## Input:
### Standards
{principle}

### Counselor Message
{consultant}

### Client Response
{client}

## Output
\end{lstlisting}

\textbf{3) Prompt for Principle-Adherence Check }
\begin{lstlisting}[numbers=none]
You are a helpful and precise assistant capable of evaluating and correcting simulated client responses.
You will receive the client's profile, speaking style, previous conversation history, the counselor's message, the simulated client's response, and a set of evaluation criteria.

Please determine whether the client's response aligns with the given criteria.
1a) Answer a set of questions to assess whether the response meets the criteria. Valid answers are: Yes, No, or Not Applicable (N/A). If you believe a question is not relevant to the current context, select "Not Applicable (N/A)."
1b) Explain your answers.

Generate a new client response.
2a) If you answered "No" to any question, write a better response that satisfies the criteria. The new response should align with the client's role description and speaking style, and match the indicated gender and age. Unless explicitly required by the criteria, do not make the response unnecessarily long or more coherent. Avoid sharing too much information at once, and keep each response under 50 characters. The new response should not be a paraphrase of the original. Instead of explicitly stating the client's emotions and feelings, the new response should express them indirectly.
2b) If you cannot generate a new response, return the original one.
2c) Provide the reason why the new response is better.

Please return the output in the following JSON format. Do not output anything except the JSON content.
```json
{
  "result": {
    "answers": [],  // List of answers to the evaluation questions
    "justification": [],  // Explanation for your answers
    "response": "",  // New client response. If there is prior conversation history, do not begin with "Hi"
    "reasoning": ""  // Explanation of why the new response is better and not a paraphrase of the original
  }
}
```

## Input:
### Questions
{question}

### Client Profile
{background}

### Client Speaking Style
{style}

### Conversation History
{history}

### Counselor Message
{consultant}

### Client Response
{client}

## Output
\end{lstlisting}

\subsection{Multidimensional Scale Evaluation}
We perform a comprehensive evaluation across three core dimensions: \textbf{therapeutic relationship, empathic understanding, and counseling skills}. Each dimension is assessed using well-established psychological scales to ensure the evaluation remains professionally rigorous. 

In this section, we introduce the three evaluation scales and describe how their scale items are further categorized to facilitate model capability analysis. For the WAI scale, we directly adopt the original item categorization. In contrast, since the BLRI and CCS-R scales do not provide predefined categories, we collaborate with multiple psychologists to develop detailed classifications, ensuring their professional validity.

\subsubsection{Therapeutic Relationship (WAI)}
The Working Alliance Inventory (WAI) measures the therapeutic relationship across three core dimensions: Counseling Goal, Task Agreement and Emotional Bond, as detailed in Table~\ref{tab:WAI}.

\begin{table}[h!]
\centering
\setlength{\tabcolsep}{3pt}
\begin{tabular}{lp{0.6\linewidth}}
\toprule
\textbf{Category} & \textbf{Items} \\
\midrule
\addlinespace
\textbf{Counseling Goal} & 
4. Counselor and I collaborate on setting goals for my therapy. \newline
6. Counselor and I are working towards mutually agreed upon goals. \newline
8. Counselor and I agree on what is important for me to work on. \newline
11. Counselor and I have established a good understanding of the kind of changes that would be good for me. \\
\addlinespace
\midrule
\addlinespace
\textbf{Task Agreement} & 
1. As a result of these sessions I am clearer as to how I might be able to change. \newline
2. What I am doing in therapy gives me new ways of looking at my problem. \newline
10. I feel that the things I do in therapy will help me to accomplish the changes that I want. \newline
12. I believe the way we are working with my problem is correct. \\
\addlinespace
\midrule
\addlinespace
\textbf{Emotional Bond} & 
3. I believe counselor likes me. \newline
5. Counselor and I respect each other. \newline
7. I feel that counselor appreciates me. \newline
9. I feel counselor cares about me even when I do things that he/she does not approve of. \\
\addlinespace
\bottomrule
\end{tabular}
\caption{Distribution of WAI Scale Items into Categories. Ratings are made on a 5-point frequency scale: 1=Seldom, 2=Sometimes, 3=Fairly Often, 4=Very Often, 5=Always.}
\label{tab:WAI}
\end{table}

\subsubsection{Empathic Understanding (BLRI)}
Empathy is assessed using the empathic understanding subscale of the Barrett-Lennard Relationship Inventory (BLRI). The "Focus-on-Empathy" subscale is a 12-item adaptation of the BLRI specifically designed for investigators who wish to focus mainly on the measure of empathic understanding.

This scale evaluates counselors’ empathic abilities across multiple dimensions, including Cognitive Empathy, Affective Empathy, Differentiated Empathy and Inner Pattern, as detailed in Table~\ref{tab:BLRI}. The BLRI includes a mix of positively and negatively worded items to reduce response bias due to rating inertia. Specifically, items 2, 5, 6, 8, 10, and 11 are reverse-scored.

\begin{table}[h!]
\centering
\setlength{\tabcolsep}{2pt}
\begin{tabular}{lp{0.53\linewidth}}
\toprule
\textbf{Category} & \textbf{Items} \\
\midrule
\addlinespace
\textbf{Cognitive Empathy} & 
3. Counselor nearly always sees exactly what I mean. \newline
5. Counselor does not understand me. \newline
7. Counselor realizes what I mean even when I have difficulty in saying it. \newline
8. Counselor doesn’t listen and pick up on what I think and feel. \newline
9. Counselor usually understands the whole of what I mean. \\
\addlinespace
\midrule
\addlinespace
\textbf{Affective Empathy} & 
1. Counselor usually senses or realizes what I am feeling. \newline
2. Counselor reacts to my words but does not see the way I feel. \newline
4. Counselor appreciates just how the things I experience feel to me. \newline
10. Counselor doesn’t realize how sensitive I am about some of the things we discuss. \\
\addlinespace
\midrule
\addlinespace
\textbf{Differentiated Empathy} & 
12. When I am hurting or upset counselor recognizes my painful feelings without becoming upset him/herself. \\
\addlinespace
\midrule
\addlinespace
\textbf{Inner Pattern} & 
6. Counselor’s own attitude toward things I do or say gets in the way of understanding me. \newline
11. Counselor’s response to me is so fixed and automatic that I don’t get through to him/her. \\
\addlinespace
\bottomrule
\end{tabular}
\caption{Distribution of BLRI Scale Items into Categories. Ratings are made on a 6-point agreement scale: -3=NO, I strongly feel that it is not true; -2=No, I feel it is not true; -1=(No) I feel that it is probably untrue, or more untrue than true; +1=(Yes) I feel that it is probably true, or more true than untrue; +2=Yes, I feel it is true; +3=YES, I strongly feel that it is true.}
\label{tab:BLRI}
\end{table}

\subsubsection{Counseling Skills (CCS-R)}
Counseling skills are evaluated using the Counselor Competencies Scale—Revised (CCS-R), including Probing Techniques, Facilitate Therapeutic Environment, Reflecting and Change Facilitation, as detailed in Table~\ref{tab:CCS-R}. The CCS-R provides counselors and trainees with direct feedback regarding their demonstrated ability to apply counseling skills, offering the counselors and trainees practical areas for improvement to support their development as effective professional counselors.

\begin{table}[h!]
\centering
\setlength{\tabcolsep}{3pt}
\begin{tabular}{p{0.38\linewidth}p{0.56\linewidth}}
\toprule
\textbf{Category} & \textbf{Items} \\
\midrule
\addlinespace
\textbf{Probing Techniques} & 
1. Includes Minimal Encouragers \& Door Openers such as ``Tell me more about...'', ``Hmm'' \newline
2. Use of Appropriate Open \& Closed Questioning (e.g., avoidance of double questions) \\
\addlinespace
\midrule
\addlinespace
\textbf{Facilitate Therapeutic Environment} & 
10. Expresses accurate empathy \& care. Counselor is ``present'' and open to client. (includes immediacy and concreteness) \newline
11. Counselor expresses appropriate respect \& unconditional positive regard \\
\addlinespace
\midrule
\addlinespace
\textbf{Reflecting} & 
3. Basic Reflection of Content – Paraphrasing \newline
4. Reflection of Feelings \newline
5. Summarizing content, feelings, behaviors, \& future plans \newline
6. Advanced Reflection of Meaning including Values and Core Beliefs (taking counseling to a deeper level) \\
\addlinespace
\midrule
\addlinespace
\textbf{Change Facilitation} & 
7. Counselor challenges client to recognize \& evaluate inconsistencies. \newline
8. Counselor collaborates with client to establish realistic, appropriate, \& attainable therapeutic goals \newline
9. Counselor focuses (or refocuses) client on his or her therapeutic goals – i.e., purposeful counseling \\
\addlinespace
\bottomrule
\end{tabular}
\caption{Distribution of CCS-R Scale Items into Categories. Ratings are made on a 5-point scale: 1=Harmful, 2=Below Expectations / Insufficient / Unacceptable, 3=Near Expectations / Developing towards Competencies, 4=Meets Expectations / Demonstrates Competencies, 5=Exceeds Expectations / Demonstrates Competencies.}
\label{tab:CCS-R}
\end{table}

\FloatBarrier

\subsection{Differential Performance Across Client Characteristics}
At the end of the main text, we analyze how the model's counseling performance varies across clients with different Big Five personality traits. In addition, we further analyze the effects of \textbf{Counseling Topic}, \textbf{Age Group}, and \textbf{Gender}, and present the results here. Significance tests are conducted for all results (*: p $<$ 0.05, **: p $<$ 0.01, ***: p $<$ 0.001).

\subsubsection{Counseling Topic}
\begin{figure}[h]
\centering
\includegraphics[width=1\columnwidth]{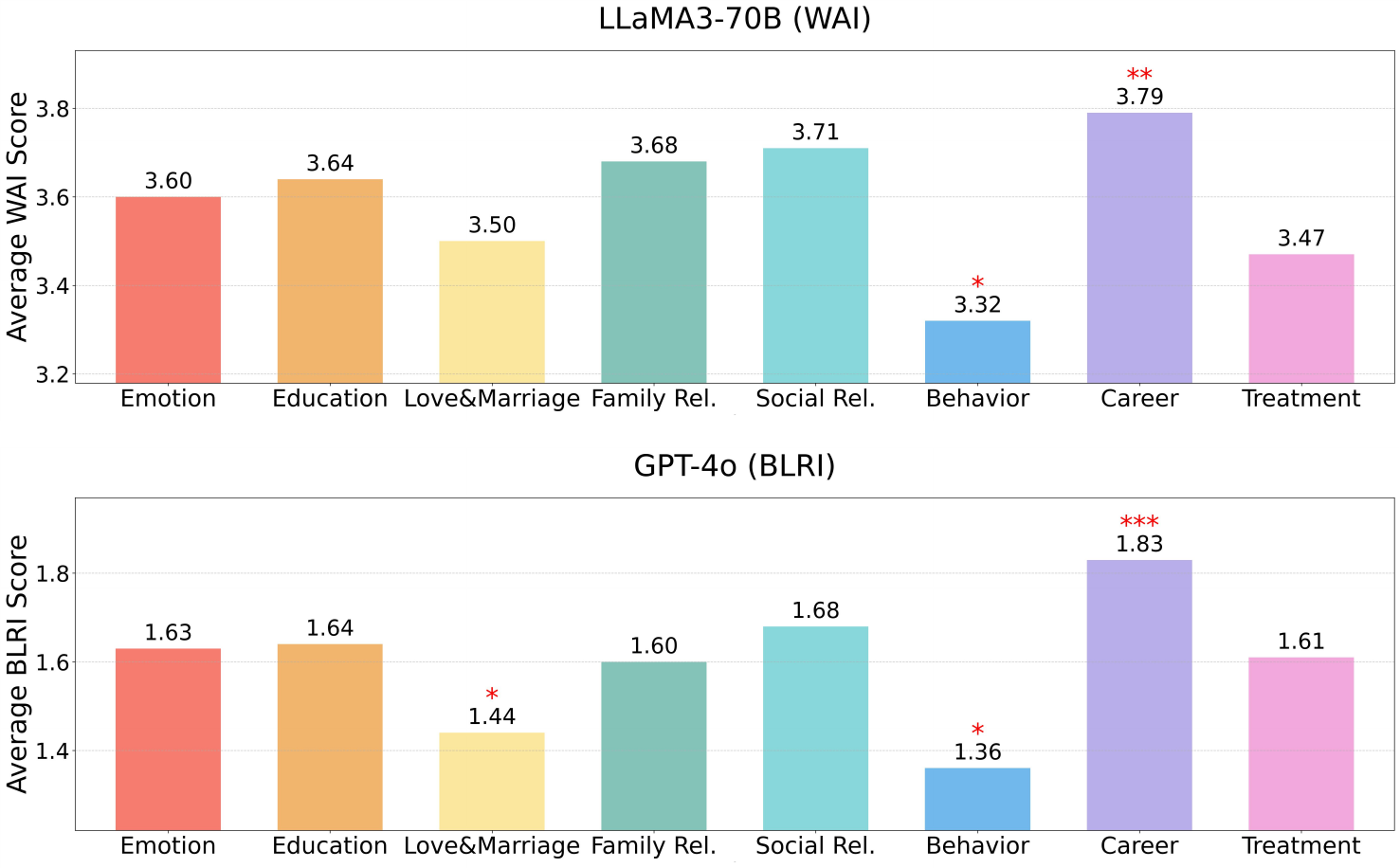} 
\caption{The generalizability results grouped by Counseling Topic. ``Rel.''stands for ``Relationship''.}
\label{subject}
\end{figure}

Statistical analysis of outcomes by counseling topics reveals that most models perform poorly on behavior problems, as shown in Figure~\ref{subject}. Their counseling style tends to be highly structured and goal-oriented, which proves effective for concrete topics such as career development that require planning and informational support. However, this approach becomes problematic when dealing with clients presenting behavioral issues characterized by high resistance, low motivation, deep emotional distress, or entrenched behavioral patterns. In such cases, the mismatch between the model’s guidance and the client’s actual needs often leads to ineffective or inappropriate recommendations.

\subsubsection{Age Group}
\begin{figure}[h]
\centering
\includegraphics[width=1\columnwidth]{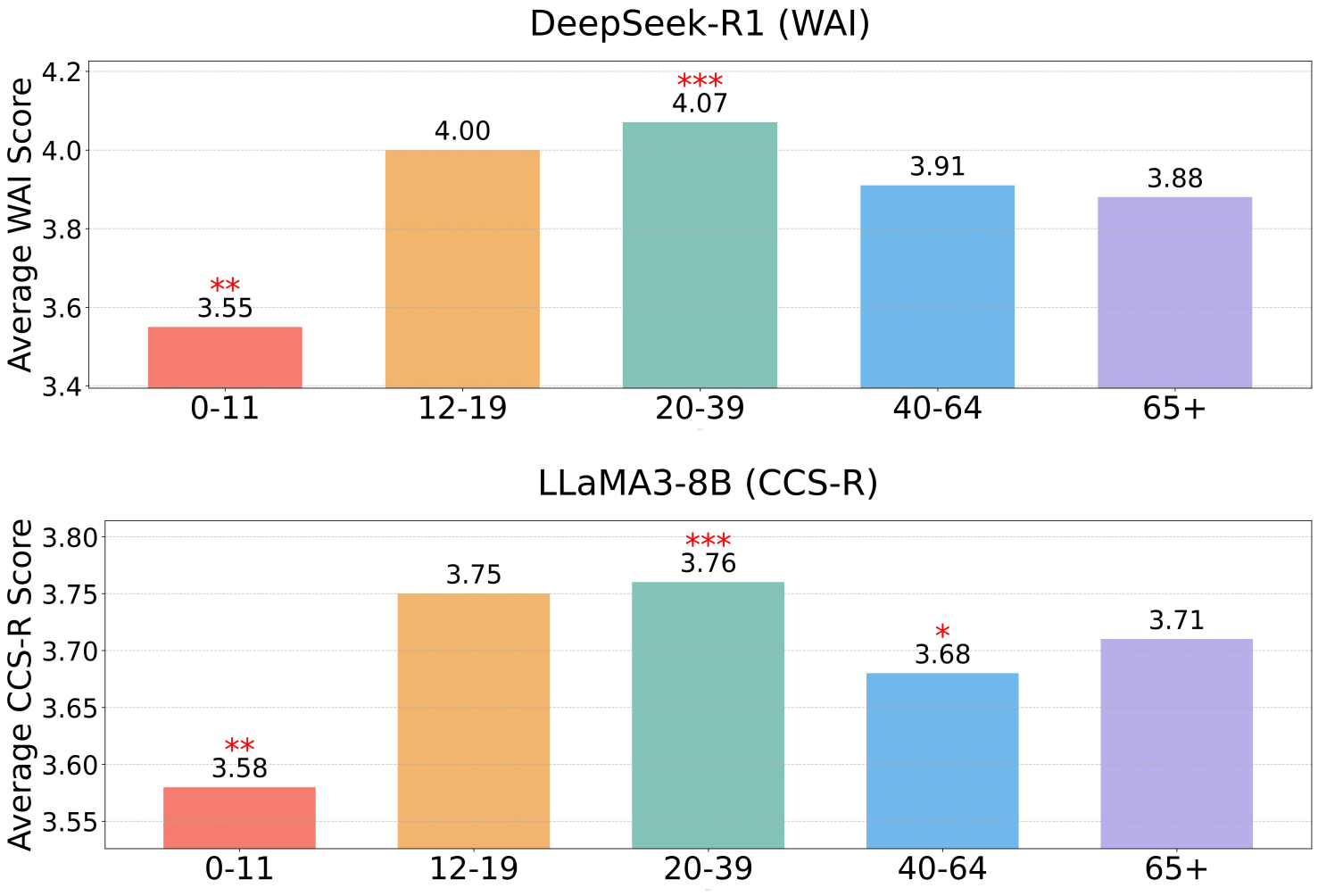} 
\caption{The generalizability results grouped by Age Group.}
\label{age}
\end{figure}

We also conduct statistical analysis by age groups, as shown in Figure~\ref{age}. 

The result shows that the 0–11 age group scores significantly lower than other age groups, while the 20–39 age group scores significantly higher. Case analyses by psychiatrists suggest that this is because the models attempt to engage young children in cognitively demanding, logic-based dialogues, such as cognitive restructuring, which exceed their receptivity. For these children, timely emotional support is more critical.

\subsubsection{Gender}
\begin{figure}[H]
\centering
\includegraphics[width=1\columnwidth]{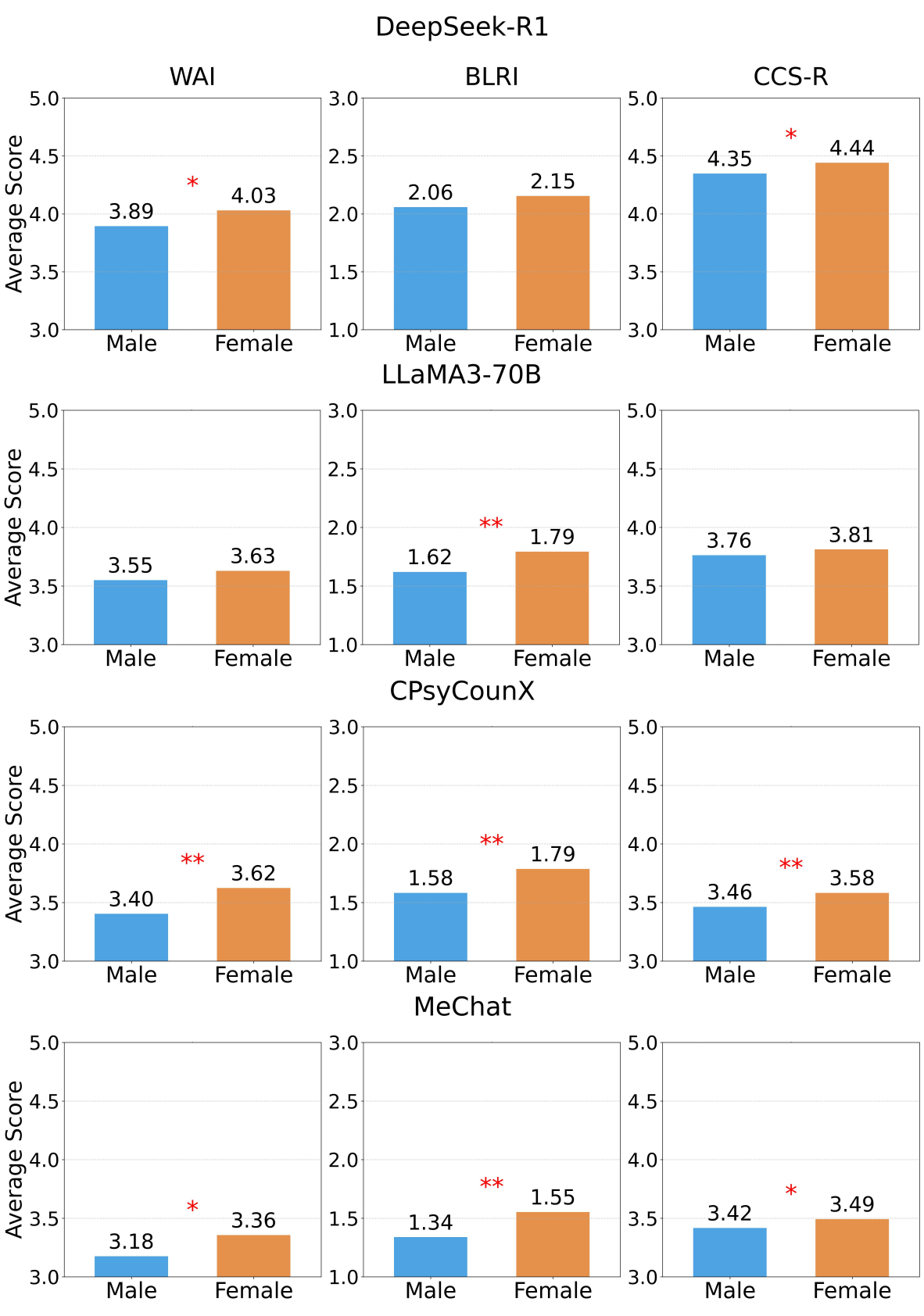} 
\caption{The generalizability results grouped by Gender.}
\label{gender}
\end{figure}

In addition, we conduct statistical analysis by gender and find that LLMs consistently achieve significantly higher counseling scores with female clients than male, as shown in Figure~\ref{gender}. Female clients' focus on emotional expression and relationships aligns well with traditional talk therapy. Conversely, male clients, who are typically more action-oriented and emotionally reserved, often resist the vulnerability required in counseling. Consequently, an approach combining emotional resonance with supportive tasks tends to build a strong therapeutic alliance with women. For men expecting direct, problem-solving outcomes, however, this strategy can feel indirect and ineffective.

Compared to the general-purpose model, this bias is more pronounced in the specialized counseling models. We hypothesize that this may result from gender imbalance in the training data used for specialization. This finding further highlights the need to prioritize gender fairness in future model development by ensuring diversity and balance in training data across genders.

\setlength{\leftmargini}{20pt}
\makeatletter\def\@listi{\leftmargin\leftmargini \topsep .5em \parsep .5em \itemsep .5em}
\def\@listii{\leftmargin\leftmarginii \labelwidth\leftmarginii \advance\labelwidth-\labelsep \topsep .4em \parsep .4em \itemsep .4em}
\def\@listiii{\leftmargin\leftmarginiii \labelwidth\leftmarginiii \advance\labelwidth-\labelsep \topsep .4em \parsep .4em \itemsep .4em}\makeatother

\setcounter{secnumdepth}{0}
\renewcommand\thesubsection{\arabic{subsection}}
\renewcommand\labelenumi{\thesubsection.\arabic{enumi}}

\newcounter{checksubsection}
\newcounter{checkitem}[checksubsection]

\newcommand{\checksubsection}[1]{%
  \refstepcounter{checksubsection}%
  \paragraph{\arabic{checksubsection}. #1}%
  \setcounter{checkitem}{0}%
}

\newcommand{\checkitem}{%
  \refstepcounter{checkitem}%
  \item[\arabic{checksubsection}.\arabic{checkitem}.]%
}
\newcommand{\question}[2]{\normalcolor\checkitem #1 #2 \color{blue}}
\newcommand{\ifyespoints}[1]{\makebox[0pt][l]{\hspace{-15pt}\normalcolor #1}}

\end{document}